\documentclass[a4paper]{article}

\usepackage{INTERSPEECH2019}
\usepackage{multirow}
\usepackage{commath}
\usepackage{amsmath,amssymb}
\usepackage{xspace}
\DeclareMathOperator{\E}{\mathbb{E}}

\title{Unsupervised End-to-End Learning of Discrete Linguistic Units for Voice Conversion}
\name{Andy T. Liu$^1$, Po-chun Hsu$^1$, Hung-yi Lee$^1$}
\address{
  $^1$College of Electrical Engineering and Computer Science, National Taiwan University}
\email{\{r07942089, r07942095, hungyilee\}@ntu.edu.tw}

\begin{document}

\maketitle

\begin{abstract}
 We present an unsupervised end-to-end training scheme where we discover discrete subword units from speech without using any labels. The discrete subword units are learned under an ASR-TTS autoencoder reconstruction setting, where an ASR-Encoder is trained to discover a set of common linguistic units given a variety of speakers, and a TTS-Decoder trained to project the discovered units back to the designated speech. We propose a discrete encoding method, Multilabel-Binary Vectors (MBV), to make the ASR-TTS autoencoder differentiable. We found that the proposed encoding method offers automatic extraction of speech content from speaker style, and is sufficient to cover full linguistic content in a given language. Therefore, the TTS-Decoder can synthesize speech with the same content as the input of ASR-Encoder but with different speaker characteristics, which achieves voice conversion (VC). We further improve the quality of VC using adversarial training, where we train a TTS-Patcher that augments the output of TTS-Decoder. Objective and subjective evaluations show that the proposed approach offers strong VC results as it eliminates speaker identity while preserving content within speech. In the ZeroSpeech 2019 Challenge, we achieved outstanding performance in terms of low bitrate.

\end{abstract}

\section{Introduction}
Despite that human speech inherently carries linguistic features that represent textual information, modern text-to-speech training pipelines still require parallel speech and text transcription pairs \cite{wang2017tacotron}\cite{shen2018natural}\cite{ping2017deep}. Parallel speech and transcripts may not always be available, and is costly to acquire, however human speech alone can be easily gathered. In this work, we focus on utilizing the advantage of unlabeled speech to discover discrete linguistic units, where machine learns to uncover the linguistic features hidden in human utterance without any supervision. We use these discovered linguistic units for voice conversion (VC) and achieved outstanding results.

Embedding audio signals into latent representations has been a well studied practice \cite{Huang2018ImprovedAE}\cite{he2016multi}\cite{Settle2016DiscriminativeAW}\cite{Chung2016AudioWU}\cite{levin2013fixed}\cite{Hsu2017UnsupervisedLO}\cite{Jansen2018UnsupervisedLO}. 
Former studies have also attempted to encode speech content into various representations for voice conversion, including the use of disentangled autoencoders \cite{chou2018multi}, VAEs \cite{hsu2016voice} or GANs \cite{gao2018voice}. 
continuous vectors are the most common approach, however when it comes to encoding speech content, they may not be the best choice. 
As they are unlike the discrete phonemes that we often used to represent human language. 
Previous works \cite{badino2014auto}\cite{Chung2015AnID}\cite{chorowski2019unsupervised} also attempt to learn discrete representations from audio, however they did not apply the learned units for VC.

In VC, the state-of-the-art approach \cite{chou2018multi} requires additional loss in the framework of GAN to guarantee the disentanglement of learned encodings, whereas the proposed approach naturally possesses the direct ability to separate speaker style from speech content. 
In \cite{van2017neural}, VC is achieved through multiple losses together with a K-way quantized embedding space and autoregressive WaveNet, on the other hand, the proposed encoding space does not require additional training constraints.

In this work, we use an ASR-TTS autoencoder to discovers discrete linguistic units from speech, without any alignment, text label, or parallel data provided. 
The ASR-Encoder learns to encode speech from different speakers to a common set of small linguistic symbols. 
These finite set of linguistic symbols are represented by Multilabel-Binary Vectors (MBV), vectors consist of arbitrary number of zeros and ones.
The proposed MBV method is differentiable, hence allowing backpropagation of gradients and end-to-end training of an autoencoder reconstruction setting. 
While the ASR-Encoder learns a many-to-one mapping from speech to discrete subword units, a TTS-Decoder learns a one-to-many mapping from discrete subword units back to speech. 
The discrete nature of MBV allows it to innately separate linguistic content from speech, removing speaker characteristics. 
Given an utterance of a source speaker, we were able to encode its speech content using the ASR-Encoder, and perform voice conversion to generate speech with the same linguistic content but style of a target speaker using the TTS-Decoder.

\begin{figure}[t]
  \centering
  \includegraphics[width=\linewidth]{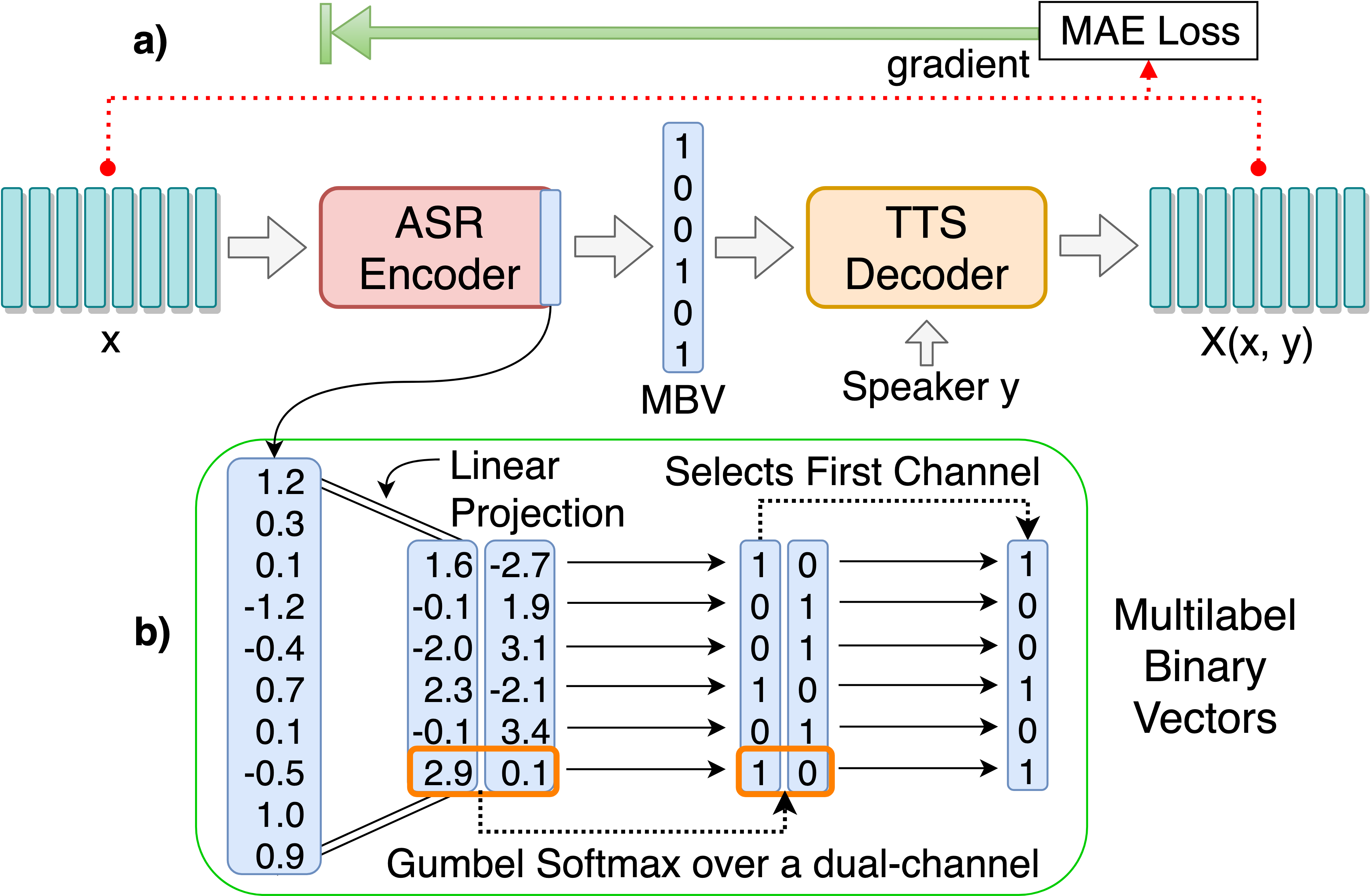}
  \caption{The ASR-TTS autoencoder framework where MBV are learned as discrete linguistic representations.}
  \label{fig:stage1}
\end{figure}

In training, the TTS-Decoder is always given an encoding from an arbitrary speaker and is trained to decode it back to the original speech of that particular speaker. During inference time, the TTS-Decoder has to take encodings of a source speaker and decode to a target speaker, an encoding-speaker pair that it never observed during training. Although speech conversion is already feasible, we propose to use additional adversarial training to compensate the training-testing inconsistency. Under the WGAN \cite{arjovsky2017wasserstein} setting, we train a TTS-Patcher in place of a generator. The TTS-Patcher learns to generate a mask that augments the output of TTS-Decoder. Furthermore, we use a target driven reconstruction loss to guide the generator's update. As a result, the quality of voice conversion performance is improved. In the ZeroSpeech 2019 Challenge \cite{zerospeech}, we achieve $2\textsuperscript{nd}\xspace$ place in terms of low bitrate under a strong dimension constraint on the Surprise Language \cite{zerospeechdata1}\cite{zerospeechdata2} leaderboard. We further show that the proposed method is capable of generating high quality intelligible speech when the dimension constraint is removed.

\begin{figure}[t]
  \centering
  \includegraphics[width=\linewidth]{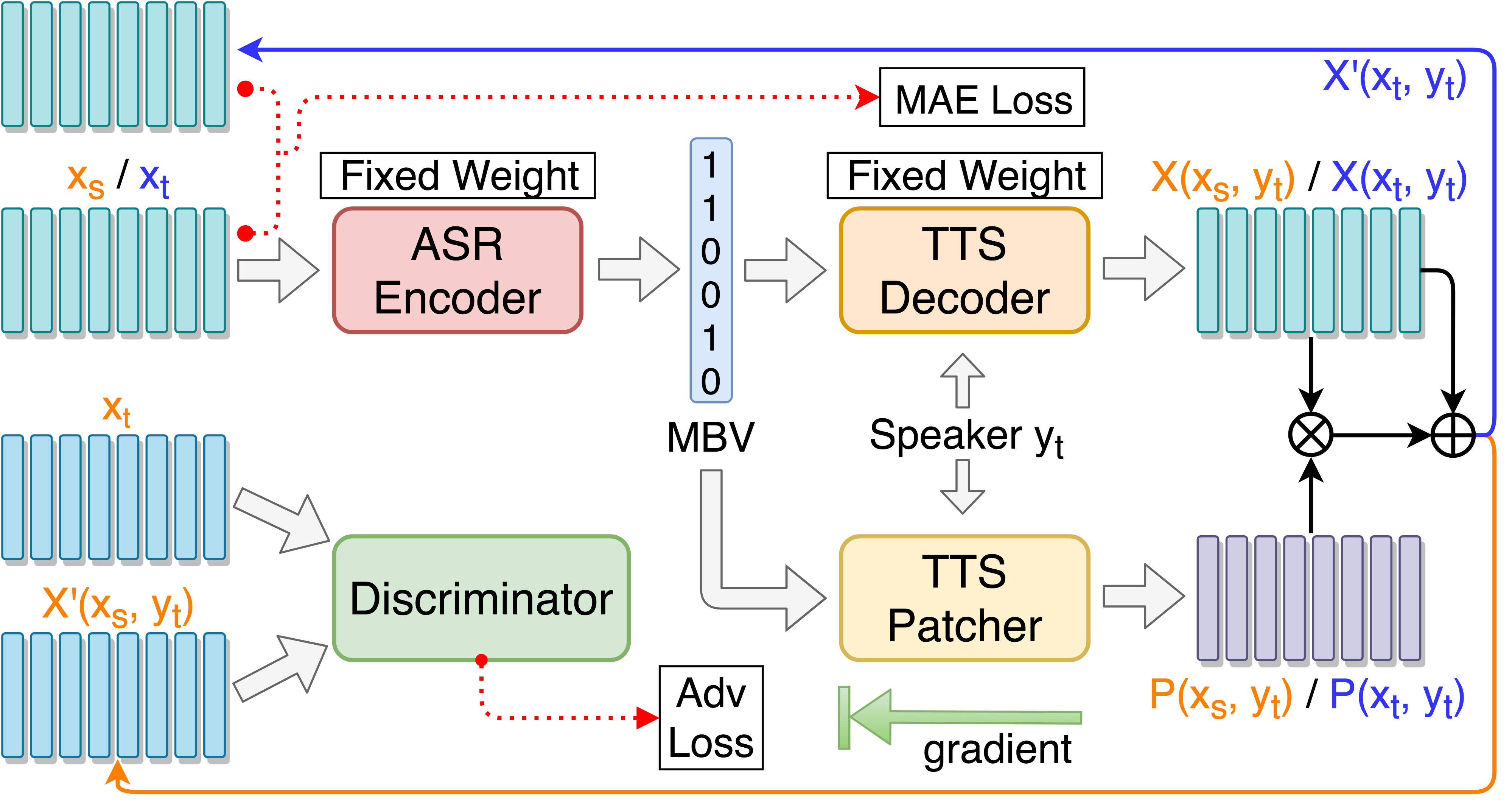}
  \caption{Target guided adversarial learning to improve voice conversion performance. We annotate the adversarial step in orange and the target guided step in blue.} 
  \label{fig:stage2}
\end{figure}

\section{Proposed method}
\subsection{Discrete linguistic units discovery}
We present an unsupervised end-to-end learning framework for distinct linguistic units discovery. 
In this stage we learn a common set of discrete encodings for all the speakers. 
Let $x \in X$ be an acoustic feature sequence where $X$ is a set of all such sequences from all source and target speakers, and $x$ is a fixed length-$r$ segment randomly sampled from $X$. 
Let $y \in Y$ be a speaker where $Y$ is the group of all source and target speakers who produce the sequence collection $X$. In a training pair $(x, y) \in (X, Y)$, the sequence $x$ is produced by speaker $y$. 

\subsubsection{Learning Multilabel-Binary Vectors}
In Figure~\ref{fig:stage1}, an ASR-TTS autoencoder framework is used to learn discrete linguistic units. 
The ASR-Encoder is trained to map input acoustic feature sequence $x \in X$ to a latent discrete encoding representation: 
\begin{equation}
    z = ASR(x),
    \label{enc_asr} 
\end{equation}
where we propose to use Multilabel-Binary Vectors (MBV) to represent $z$ generated in (\ref{enc_asr}), the discrete encoding $z$ is designed to represents the linguistic content of input speech $x$. 
We define MBV as:
\begin{equation}
    z = [e_1, e_2, \cdots, e_n] \in \mathbb{R}^n, e_i \in \{0,1\}
    \label{mbv}
\end{equation}
where $z$ is a $n$ dimension binary vector consists of arbitrary number of zeros and ones. 
The proposed MBV encoding method is differentiable, and allows end-to-end training with direct gradient backpropagation in an autoencoder framework. 
To obtain the binarized differentiable vector $z$, we linearly project a continuous output vector into a $\mathbb{R}^{n \times 2}$ space, where $n$ is the dimension of the wanted MBV. The dual-channel projection allows each dimension in a MBV to symbolize an arbitrary attribute's presence. We then perform categorical reparameterization trick with Gumbel-Softmax \cite{jang2016categorical} on the $2\textsuperscript{nd}\xspace$-dimension dual-channel, which is equivalent to asking the model to predict whether an attribute is observed in a given input.
Since the two channels are linked together by Gumbel-Softmax, simply picking one of them is sufficient, hence we select the first channel as our MBV encoding $z$.
We illustrate the process of MBV encoding in Figure~\ref{fig:stage1} (b). 
Former approaches also use Gumbel-Softmax on the output layer to obtain one-hot vectors~\cite{jang2016categorical}.
However, the resulting one-hot vectors is insufficient to represent content in speech, as the one-hot encoding space is too sparse for machines to learn linguistic meanings, which we verified in our experiments.

\subsubsection{Voice reconstruction and conversion}
Given ASR-Encoder's output, the TTS-Decoder is trained to generate an output acoustic feature sequence defined as: 
\begin{equation}
     X(x,y) = TTS(ASR(x), y), 
     \label{x'_tts} 
\end{equation}
where $X(x,y)$ is a reconstruction of $x$ from $z=ASR(x)$ given the speaker identity $y \in Y$.
Mean Absolute Error (MAE) is employed to evaluate the ASR-TTS autoencoder reconstruction loss, as MAE is reported to be able to generate sharper outputs than Mean Square Error \cite{Isola_2017_CVPR}. 
The reconstruction loss is given by:
\begin{equation}
    L_{rec}(\theta_{asr}, \theta_{tts}) = E_{(x, y) \sim (X, Y)} \norm{x - X(x,y)}_1,       
    \label{L_rec}
\end{equation}
where $\theta_{asr}$ and $\theta_{tts}$ are the parameters of the ASR-Encoder and TTS-Decoder, respectively. 
We uniformly sample $(x, y)$ for training in (\ref{L_rec}). 
Because the speaker identity is provided to the TTS-Decoder, the proposed MBV encodings $z$ is able to learn an abstract space that is invariant to speaker identity and only encodes the content of speech, without using any form of linguistic supervision. 

At inference time, given a source speech $x_s$, and target speaker $y_t$, the TTS-Decoder can generate the voice of the target speaker $y_t$ using the linguistic content $ASR(x_s)$ from $x_s$:
\begin{equation}
    X(x_s, y_t) = TTS(ASR(x_s), y_t),
    \label{X()}
\end{equation}
$X(x_s, y_t)$ is the output of TTS-Decoder, which has the linguistic content of $x_s$ but style of speaker $y_t$.

\subsection{Target guided adversarial learning}
With the learned speaker invariant ASR-Encoder mapping in (\ref{enc_asr}), we successfully represent speech content with discrete binary vectors MBVs. In this section, we describe how adversarial training is used to boost VC quality based on the discrete linguistic units learned in Section 2.1. We train a TTS-Patcher in an unsupervised manner, in which the TTS-Patcher generates a spectrum mask that residually augments the output of equation (\ref{X()}), resulting in a more precise voice conversion result.

We define two sets of speakers, source speaker set and target speaker set, where we aim to convert the speech of a source speaker into a target speaker's style while preserving its content. Let $x_s \in X_S$ and $x_t \in X_T$ be sequences where $X_S$ and $X_T$ are the set of sequences from source speakers and target speakers, respectively. Let $y_s \in Y_S$ be a speaker from the set of all source speakers $Y_S$ who produce $X_S$, and let $y_t \in Y_T$ be a speaker from the set of all target speakers $Y_T$ who produce $X_T$. 

\subsubsection{Adversarial learning step}
In Figure~\ref{fig:stage2}, a TTS-Patcher is trained as a generator under an adversarial learning setting. 
The TTS-Patcher takes $ASR(x_s)$ and $y_t$ as input, and generates a spectrum mask $P(x_s, y_t)$ that ranges from zero to one, and modifies $X(x_s, y_t)$ through a residual augmentation:
\begin{equation}
    X'(x_s, y_t) = ( P(x_s, y_t) \otimes X(x_s, y_t) ) \oplus X(x_s, y_t). 
    \label{X'()}
\end{equation}
The $\oplus$ and $\otimes$ symbols indicate element-wise addition and multiplication, respectively. 
In (\ref{X'()}), $x_s$ is a randomly sampled input speech segment from source speaker $y_s$, where $y_t$ is a randomly sampled target speaker. 
$X(x_s, y_t)$ is the voice conversion utterance obtained in (\ref{X()}), $P(x_s, y_t)$ is the output of TTS-Patcher, and finally $X'(x_s, y_t)$ the augmented spectrum.

A discriminator $D$ is trained to distinguish whether an input acoustic feature sequence is real or reconstructed by machine.  Since naive GAN \cite{NIPS2014_5423} is notoriously hard to train, we minimize the Wasserstein-1 distance between real and fake distributions instead, as proposed in the WGAN \cite{arjovsky2017wasserstein} formulation:
\begin{equation}
\begin{aligned}
    L_{WGAN} &= \E_{x_t \sim X_T}[D(x_t)] \\
    &- \E_{x_s \sim X_S, y_t \sim Y_T}[D(X'(x_s, y_t))].
    \label{eq6}
\end{aligned}
\end{equation}
The discriminator computes the Wasserstein-1 distance of two distributions: real data $x_t$ sampled from the target speaker set $X_T$, and augmented voice conversion outputs from (\ref{X'()}). We use the alternative WGAN-GP \cite{gulrajani2017improved} to enforce the 1-Lipschitz constraint required by $D$, where weight clipping is replaced with gradient penalty. On the last layer of the discriminator, we stretch an additional layer that learns a classifier to predict speaker from a given speech. 
This allows the discriminator to consider input spectrum's fidelity and speaker identity at the same time \cite{odena2017conditional}.

\subsubsection{Target guided training step}
The decoupled learning \cite{zhang2018decoupled} of ASR-TTS autoencoder and TTS-Patcher stabilizes the GAN \cite{NIPS2014_5423} training process. However we found that under the adversarial learning scheme, the TTS-Patcher can easily learn to deceive the discriminator by over-adding style, this greatly compromises the original speech content. This is caused by the discriminator's inability to discriminate utterances with incorrect or ambiguous content, the discriminator only learns to focus on speaker style. As a solution, we propose to add an additional target guided training step, we apply additional reconstruction loss after every adversarial step, as shown in Figure~\ref{fig:stage2}. Instead of converting $x_s$, the ASR-Encoder now takes a segment of target speech $x_t$ as input, equation (\ref{X'()}) then becomes:
\begin{equation}
     X'(x_t, y_t) = ( P(x_t, y_t) \otimes X(x_t, y_t) ) \oplus X(x_t, y_t), 
    \label{X'()-2}
\end{equation}
and we minimize MAE between $x_t$ and $X'(x_t, y_t)$:
\begin{equation}
    L_{rec}(\theta_{p}) = E_{(x_t, y_t) \sim (X_T, Y_T)} \norm{x_t - X'(x_t, y_t)}_1,
    \label{L_rec-2}
\end{equation}
where $\theta_{p}$ is the parameter of the TTS-Patcher. This loss effectively guides the TTS-Patcher's update under adversarial settings, as the added style is regularized to preserve intelligibility.

\section{Implementation}
The ASR-Encoder is inspired by the CBHG module \cite{wang2017tacotron}, where the linear output of ASR-Encoder is fed to the MBV encoding module. We add noise in training by adding dropout layers in the ASR-Encoder as suggested in \cite{isola2017image}. The TTS-Decoder and TTS-Patcher have identical model architectures, where we use pixel shuffle layers to generate high resolution spectrum \cite{shi2016real}. We add speaker embedding on the feature map of all layers, where a distinct embedding is learned for all different layers as different information may be needed for each layer. The discriminator is consist of 2D-convolution blocks for temporal texture capturing, and convolutions projection layers followed by fully-connected output layers. 
We trained the network using Adam \cite{kingma2014adam} optimizer and a batch size of 16. In the discrete linguistic units discovery stage, we train the ASR-TTS autoencoder for 200k mini-batches. In the target guided adversarial learning stage we train the model for 50k mini-batches, in one batch we train a step of adversarial learning including 5 iterations of discriminator update and 1 iteration of generator update, followed by a target guided reconstruction step. 

We train and evaluate our model using the ZeroSpeech 2019 English dataset \cite{zerospeech}. In particular, we use the ``Train Unit Dataset'' as our source speaker set $X_S$, the ``Train Voice Dataset'' as our target speaker set $X_T$, and we evaluate models with the ``Test Dataset''.
We used log-magnitude spectrograms as acoustic features, the detailed settings are in Table~\ref{tb:preprocessing}. During training, our model is trained to process 128 consecutive overlapping frames of spectrogram, where we uniformly sample from the dataset. At inference time, for a given input with more than 128 frames, the model process them as segments and concatenate the outputs on the time-axis. Source code are publicly available\footnote{https://github.com/andi611/ZeroSpeech-TTS-without-T}. 

\begin{table}[th]
\caption{Acoustic feature preprocess settings}
  \label{tb:preprocessing}
  \centering
\begin{tabular}{c | c || c | c}
    Pre-emphasis & $0.97$ & Sample rate & $16$k\\
    \hline
    Frame length & $50$ ms & Mel-spec & $80$ \\
    \hline
    Frame shift & $12.5$ ms & Linear-spec & $1024$ \\
    \hline
    Window type & Hann & Vocoder & Griffin-Lim \\
\end{tabular}
\end{table}

\begin{table}[t]
\caption{Comparison of different latent representations.}
  \label{tb:disentanglement}
  \centering
\begin{tabular}{c | ll}
    \toprule
    \textbf{Types of encodings} & \textbf{Dim} & \textbf{Acc} \\
    \midrule
    One-hot & 1024 & 43.3\% \\
    continuous & 1024 & 84.1\% \\
    continuous & 128 & 79.9\% \\
    continuous (with add'l loss) & 1024 & 78\% \\
    continuous (with add'l loss) & 128 & 81.3\% \\
    Ours (MBV) & 1024 &  \textbf{92.3}\% \\
    Ours (MBV) & 128 & \textbf{93.9\%} \\
    \bottomrule
\end{tabular}
\end{table}

\section{Experiments}
We compare the proposed MBV encodings with one-hot encodings, continuous encodings, and continuous encodings with additional loss \cite{chou2018multi}, all of which under the same autoencoder training setting as described in Section 3.

\subsection{Degree of disentanglement}
To evaluate the degree of disentanglement for the proposed MBV with respect to speaker characteristics, we trained a speaker verification classifier that takes spectrum as input and predicts speaker identity. 
We encode speech from source speaker and convert it to a target speaker. 
With the pre-trained classifier we measure target speaker classification accuracy on the converted results. 
A disentangled representation should produce voice similar to the target speakers and leads to higher classification accuracy. 
The results are shown in Table~\ref{tb:disentanglement}, where the $Dim$ column indicates the dimension $n$ of encoding vectors $z \in \mathbb{R}^n$. 
One-hot encodings are insufficient to encode speech, resulting in poor conversion results. 
continuous encodings are incapable of disentangling content from style, resulting in lower classification accuracy. 
Where as the proposed MBV encodings has the ability to preserve speech content while removing speaker style.
Although both one-hot vectors and MBV are discrete, each dimension of one-hot vector corresponds to a linguistic unit, while each dimension of MBV may corresponds to a pronunciation attribute. This makes MBV more data efficient than one-hot vectors.

\subsection{Subjective and objective evaluation}
We perform subjective human evaluation on the converted voices. We use 20 subjects to grade each method on a 1 to 5 scale under two measures: the naturalness of speech and the similarity in speaker characteristics to the target speaker.
In Table~\ref{tb:human_eval} we show the result of our evaluation, the proposed method results in significant increase of target similarity with a slight degrade of naturalness.
We easily achieve comparable speech intelligibility as ordinary continuous methods, while achieving better voice conversion quality with more disentanglement (Table~\ref{tb:disentanglement}). Subjective and objective evaluations suggest that the proposed MBV method eliminates speaker identity while reserving content within speech, and is suitable for voice conversion.

\begin{table}[th]
\caption{Results of subjective human evaluation. All methods used an encoding dimension of 1024 if not specified otherwise.}
  \label{tb:human_eval}
  \centering
\begin{tabular}{l | ll}
    \toprule
    \textbf{Types of encodings} & \textbf{naturalness} & \textbf{similarity} \\
    \midrule
    continuous & 3.80 & 2.14 \\
    continuous (with add'l loss) & 3.21 & 2.58 \\
    Ours (MBV with dim 6) & 1.61 & 1.51 \\
    Ours (MBV) & 3.36 & \textbf{3.06} \\
    Ours (with adv. training) & 2.57 & \textbf{3.15} \\
    \bottomrule
\end{tabular}
\end{table}

\subsection{Encoding dimension analysis}
We use several objective measures to determine the quality of an encoding, these measurements are shown in Table~\ref{tb:dim}. 
The $CER$ column is the output Character Error Rate (CER) from a pre-trained ASR, where we use the ASR results of real input voice as ground truth, which measures intelligibility of the converted speech.
The $BR$ column measures the bitrate (amount of information) that encodings carry in average with respect to the testing set, as suggested in \cite{zerospeech}. The $ABX$ column measures the machine ABX score, which indicates the goodness of encoding quality \cite{schatz:hal-00918599}\cite{zerospeech}. The $distinct$ column indicates the number of unique symbols used to encode speech in the test set. Lower values suggest a better performance for all the measures described above. In Table~\ref{tb:dim}, we compare the proposed method with other approaches along side with the baseline model \cite{ondel2016variational}\cite{wu2016merlin} demonstrated in \cite{zerospeech}. When compared to other approaches, the proposed method achieves lower $BR$ and $distinct$ values with comparable $ABX$ scores.
Due to the discrete and differentiable nature of MBV, the proposed method can be used in other unsupervised end-to-end clustering or classification tasks, where other approaches may fail to generalize.

\begin{table}[t]
  \caption{Performance of different encoding dimensions.}
  \label{tb:dim}
  \centering
  \begin{tabular}{ l|l|llll }
    \toprule
    \textbf{Method} & \textbf{Dim} & \textbf{CER} & \textbf{BR} & \textbf{ABX} & \textbf{distinct}               \\
    \midrule
    Baseline & 200 & 1.000 & 71.98 & 35.90 & 65\\
    \midrule
    \multirow{2}{2em}{Cont.} & 1024 & 0.036 & 138.45 & 31.83 & 16849\\
    & 128 & 0.040 & 138.45 & 33.96 & 16849\\
    \midrule
    \multirow{11}{2em}{Ours} & 1024 & 0.196 & 138.45 & 32.02 & 16849\\
    & 512 & 0.313 & 138.45 & 32.82 & 16849\\
    & 256 & 0.430 & 138.45 & 32.52 & 16849\\
    & 128 & 0.629 & 138.45 & 31.58 & 16849\\
    & 64 & 0.717 & 138.35 & 32.57 & 16772\\
    & 32 & 0.797 & 134.80 & 31.82 & 14591\\
    & 16 & 0.887 & 105.96 & 35.62 & 3723\\
    & 8 & 0.998 & 61.79 & 38.10 & 146\\
    & 7 & 0.998 & 55.97 & 37.71 & 94\\
    & 6 & 1.000 & \textbf{48.78} & 39.60 & \textbf{51}\\
    & 5 & 1.000 & \textbf{41.32} & 41.79 & \textbf{28}\\
    \bottomrule
  \end{tabular}
\end{table}

\begin{figure}[t]
  \centering
  \includegraphics[width=\linewidth]{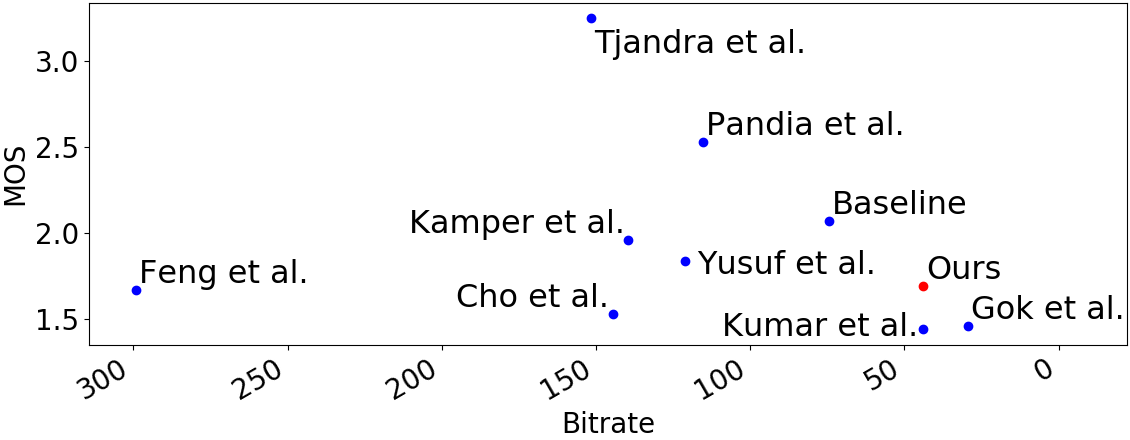}
  \caption{Partial results of the ZeroSpeech Challenge.}
  \label{fig:leaderboard}
\end{figure}

\subsection{The zero resource speech challenge competition}
We compete with other teams in the ZeroSpeech 2019 Challenge~\cite{zerospeech} at a global scale. We use the proposed method with a dimension of $6$ to achieve extremely low bitrate, and were able to encode a whole language with less than 64 distinct units, human evaluation also suggests that the produced speech are still acceptable (Table~\ref{tb:human_eval}). On the Surprise dataset \cite{zerospeechdata1}\cite{zerospeechdata2} leaderboard, the proposed method is $2\textsuperscript{nd}\xspace$ place in terms of low bitrate, while achieving higher Mean Opinion Score (MOS) and lower CER than the $1\textsuperscript{st}\xspace$ place team, as shown in Figure~\ref{fig:leaderboard}.
Although the proposed approach does not achieve high MOS because it is an inevitable trade-off with extremely low bitrate, in Table~\ref{tb:human_eval} we have shown that with a larger encoding dimension we were able to generate outstanding voice converted speech.

\section{Conclusions}
We proposed to use multilabel-binary vectors to represent the content of human speech, as its discrete nature offers a strong extraction of speaker-independent representation. We show that these discrete units naturally possess the ability of disentangling speech content and style, which makes them extremely suitable for voice conversion tasks. Also, we show that these discrete units indeed produce better style disentanglement than ordinary settings, and finally we were able to improve voice conversion results through the addition of residual augmented signals.

\bibliographystyle{IEEEtran}

\bibliography{mybib}

\end{document}